# Modeling Transportation Routines using Hybrid Dynamic Mixed Networks

**Vibhav Gogate, Rina Dechter, Bozhena Bidyuk**
Donald Bren School of Information and Computer Science
University of California, Irvine, CA 92967
{vgogate,dechter,bbidyuk}@ics.uci.edu

**Craig Rindt and James Marca**
Institute of Transportation science
University of California, Irvine, CA 92967
{jmarca,crindt}@translab.its.uci.edu

## Abstract

This paper describes a general framework called Hybrid Dynamic Mixed Networks (HDMNs) which are Hybrid Dynamic Bayesian Networks that allow representation of discrete deterministic information in the form of constraints. We propose approximate inference algorithms that integrate and adjust well known algorithmic principles such as Generalized Belief Propagation, Rao-Blackwellised Particle Filtering and Constraint Propagation to address the complexity of modeling and reasoning in HDMNs. We use this framework to model a person's travel activity over time and to predict destination and routes given the current location. We present a preliminary empirical evaluation demonstrating the effectiveness of our modeling framework and algorithms using several variants of the activity model.

## 1 INTRODUCTION

Modeling sequential real-life domains often requires the ability to represent both probabilistic and deterministic information. Hybrid Dynamic Bayesian Networks (HDBNs) were recently proposed for modeling such phenomena [Lerner, 2002]. In essence, these are factored representation of Markov processes that allow discrete and continuous variables. Since they are designed to express uncertain information they represent constraints as probabilistic entities which may have negative computational consequences. To address this problem [Dechter and Mateescu, 2004, Larkin and Dechter, 2003] introduced the framework of Mixed Networks. In this paper we extend the Mixed Networks framework to dynamic environments, allow continuous Gaussian variables, yielding Hybrid Dynamic Mixed Networks (HDMN). We address the algorithmic issues that emerge from this extension and demonstrate the potential of our approach on a complex dynamic domain of a person's transportation routines.

Focusing on algorithmic issues, the most popular approximate query processing algorithms for dynamic networks are Expectation propagation(EP) [Heskes and Zoeter, 2002] and Rao-Blackwellised Particle Filtering (RBPF) [Doucet et al., 2000]. We therefore extend these algorithms to accommodate and exploit discrete constraints in the presence of continuous probabilistic functions. Extending Expectation Propagation to handle constraints is easy, extension to continuous variables is a little more intricate but still straightforward. The presence of constraints introduces a principles challenge for Sequential Importance Sampling algorithms, however. Indeed the main algorithmic contribution of this paper in presenting a class of Rao-Blackwellised Particle Filtering algorithm, IJGP-RBPF for HDMNs which integrates a Generalized Belief Propagation component with a Rao-Blackwellised Particle Filtering scheme.

Our motivation for developing HDMNs as a modeling framework is a range of problems in the transportation literature that depend upon reliable estimates of the prevailing demand for travel over various time scales. At one end of this range, there is a pressing need for accurate and complete estimation of the global origins and destinations (O-D) matrix at any given time for an entire urban area. Such estimates are used in both urban planning applications [Sherali et al., 2003] and integrated traffic control systems based upon dynamic traffic assignment techniques [Peeta and Zilaskopoulos, 2001]. Even the most advanced techniques, however, are hamstrung by their reliance upon out-dated, pencil-and-paper travel surveys and sparsely distributed detectors in the transportation system. We view the increasing proliferation of powerful mobile computing devices as an opportunity to remedy this situation. If even a small sample of the traveling public agreed to collect their travel data and make that data publicly available, transportation management systems could significantly improve their operational efficiency. At the other end of the spectrum, *personal traffic assistants* running on the mobile devices could help travelers replan their travel when the routes they typically use are impacted by failures in the system arising from accidents or natural disasters. A common starting point for these problems is to develop an efficient formulation for learning and inferring individual traveler routines like traveler's destination and his route to destination from raw data points.

The rest of the paper is organized as follows. In the next section, we discuss preliminaries and introduce our modeling framework. We then describe two approximate inference algorithms for processing HDMN queries: an Expectation Propagation type and a Particle Filtering type. Subsequently, we describe the transportation modeling approach and present preliminary empirical results on how effectively a model is learnt and how accurately its predictions are given several models and a few variants of the relevant algorithms.

We view the contribution of this paper in addressing a complex and highly relevant real life domain using a general framework and domain independent algorithms, thus allowing systematic study of modeling, learning and inference in a non-trivial setting.

## 2 PRELIMINARIES AND DEFINITIONS

*Hybrid Bayesian Networks (HBN)* [Lauritzen, 1992] are graphical models defined by a tuple $\mathcal{B} = (X, G, P)$, where $X$ is the set of variables partitioned into discrete and continuous ones $X = \Gamma \bigcup \Delta$, respectively, $G$ is a directed acyclic graph whose nodes corresponds to the variables. $P = \{P_1, ..., P_n\}$ is a set of conditional probability distributions (CPDs). Given variable $x_i$ and its parents in the graph $pa(x_i)$, $P_i = P(x_i|pa(x_i))$. The graph structure $G$ is restricted in that continuous variables cannot have discrete variables as their child nodes. The conditional distribution of continuous variables are given by a linear Gaussian model: $P(x_i|I = i, Z = z) = N(\alpha(i) + \beta(i) * z, \gamma(i))x_i \in \Gamma$ where $Z$ and $I$ are the set of continuous and discrete parents of $x_i$, respectively and $N(\mu, \sigma)$ is a multi-variate normal distribution. The network represents a joint distribution over all its variables given by a product of all its CPDs.

A *Constraint Network* [Dechter, 2003] is a graphical model $\mathcal{R} = (X, D, C)$, where $X = \{x_1, ..., x_n\}$ is the set of variables, $D = \{D_1, ..., D_n\}$ is their respective discrete domains and $C = \{C_1, C_2, ..., C_m\}$ is the set of constraints. Each constraint $C_i$ is a relation $R_i$ defined over a subset of the variables $S_i \subseteq X$ and denotes the combination of values that can be assigned simultaneously. A *Solution* is an assignment of values to all the variables such that no constraint is violated. The primary query is to decide if the constraint network is consistent and if so find one or all solutions.

The recently proposed Mixed Network framework [Dechter and Mateescu, 2004] for augmenting Bayesian Networks with constraints, can immediately be applied to HBNs yielding the *Hybrid Mixed Networks (HMNs)*. Formally, given a HBN $\mathcal{B} = (X, G, P)$ that expresses the joint probability $P_\mathcal{B}$ and given a constraint network $\mathcal{R} = (X, D, C)$ that expresses a set of solutions $\rho$, an HMN is a pair $\mathcal{M} = (\mathcal{B}, \mathcal{R})$. The discrete variables and their domains are shared by $\mathcal{B}$ and $\mathcal{R}$ and the relationships are those expressed in $P$ and $C$. We assume that $\mathcal{R}$ is consistent. The mixed network $\mathcal{M} = (\mathcal{B}, \mathcal{R})$ represents the conditional probability $P_\mathcal{M}(x) = P_\mathcal{B}(x|x \in \rho)$ *if* $x \in \rho$ and 0 otherwise.

Dynamic Bayesian Networks are Markov models whose state-space and transition functions are expressed in a factored form using Bayesian Networks. They are defined by a prior $P(X_0)$ and a state transition function $P(X_{t+1}|X_t)$. Hybrid Dynamic Bayesian Networks (HDBNs) allow continuous variables while Hybrid Dynamic Mixed Networks (HDMNs) also permit explicit discrete constraints.

DEFINITION **2.1** *A* **Hybrid Dynamic Mixed Network (HDMN)** *is a pair* $(M_0, M_\rightarrow)$, *defined over a set of variables* $X = \{x_1, ..., x_n\}$, *where $M_0$ is an HMN defined over $X$ representing $P(X_0)$. $M_\rightarrow$ is a 2-slice network defining the stochastic process $P(X_{t+1}|X_t)$. The* 2-time-slice Hybrid Mixed network (2-THMN) *is an HMN defined over $X' \cup X''$ such that $X'$ and $X''$ are identical to $X$. The acyclic graph of the probabilistic portion is restricted so that nodes in $X'$ are root nodes and have no CPDs associated with them. The constraints are defined the usual way. The 2-THMN represents a conditional distribution $P(X''|X')$.*

The semantics of any dynamic network can be understood by unrolling the network to $T$ time-slices. Namely, $P(X_{0:t}) = P(X_0) * \prod_{t=1}^{T} P(X_t|X_{t-1})$ where each probabilistic component can be factored in the usual way, yielding a regular HMN over $T$ copies of the state variables.

The most common task over Dynamic Probabilistic Networks is filtering and prediction Filtering is the task of determining the belief state $P(X_t|e_{0:t})$ where $X_t$ is the set of variables at time $t$ and $e_{0:t}$ are the observations accumulated at time-slices 0 to $t$. Filtering can be accomplished in principle by unrolling the dynamic model and using any state-of-the art exact or approximate reasoning algorithm. The join-tree-clustering algorithm is the most commonly used algorithm for exact inference in Bayesian networks. It partitions the CPDs and constraints into clusters that interact in a tree-like manner (the join-tree) and applies message-passing between clusters. The complexity of the algorithm is exponential in a parameter called treewidth, which is the maximum number of discrete variables in a cluster. However, the stochastic nature of Dynamic Networks restricts the applicability of join-tree clustering considerably. In the discrete case the temporal structure implies tree-width which equals to the number of state variables that are connected with the next time-slice, thus making the factored representation ineffective. Even worse, when both continuous and discrete variables are present the effective treewidth is $O(T)$ when $T$ is the number of time slices, thus making exact inference infeasible. Therefore the applicable approximate inference algorithms for Hybrid Dynamic Networks are either sampling-based such as Particle Filtering or propagation-based such as Expectation Propagation. In the next two sections, we will extend these algorithms to HDMNs.

## 3 EXPECTATION PROPAGATION

In this section we extend an approximate inference algorithm called Expectation Propagation (EP) [Heskes and Zoeter, 2002] from HDBNs to HDMNs. The idea in EP (forward pass) is to perform Belief Propagation by passing messages between slices $t$ and $t+1$ along the ordering $t = 0$ to $T$. EP can be thought of as an extension of Generalized Belief Propagation (GBP) to HDBNs [Heskes and Zoeter, 2002]. For simplicity of exposition, we will extend a GBP algorithm called Iterative Join graph propagation [Dechter et al., 2002] to HDMNs and call our technique IJGP(i)-S where "S" denotes that the process is sequential. The extension is rather straightforward and can be easily derived by integrating the results in [Murphy, 2002, Dechter et al., 2002, Lauritzen, 1992, Larkin and Dechter, 2003].

IJGP [Dechter et al., 2002] is a Generalized Belief Propagation algorithm which performs message passing on a join-graph. A join-graph is collection of cliques or clusters such that the interaction between the clusters is captured by a graph. Each clique in a join-graph contains a subset of variables from the graphical model. IJGP(i) is a parameterized algorithm which operates on a join-graph which has less than $i+1$ discrete variables in each clique. The complexity of IJGP(i) is bounded exponentially by $i$, also called the $i$-bound. In the message-passing step of IJGP(i), a message is sent between any two nodes that are neighbors of each other in the join-graph. A message sent by node $N_i$ to $N_j$ is constructed by multiplying all the functions and messages in a node (except the message received from $N_j$) and marginalizing on the common variables between $N_j$ and $N_i$ (see [Dechter et al., 2002]).

IJGP(i) can be easily adapted to HDMNs (which we call IJGP(i)-S) and we describe some technical details here rather than a complete derivation due to lack of space. Note that because we are performing online inference, we need to construct the join-graph used by IJGP(i)-S in an online manner rather than recomputing the join-graph every time new evidence arrives. Murphy [Murphy, 2002] describes a method to compute a join-tree in an online manner by pasting together join-trees of individual time-slices using special cliques called the interface. [Dechter et al., 2002] describe a method to compute join-graphs from join-trees. The two methods can be combined in a straightforward way to come up with an online procedure for constructing a join-graph. In this procedure, we split the interface into smaller cliques such that the new cliques have less than $i+1$ variables. This construction procedure is shown in Figure 1.

Message-passing is then performed in a sequential way as follows. At each time-slice $t$, we perform message-passing over nodes in $t$ and the interface of $t$ with $t-1$ and $t+1$ (shown by the ovals in Figure 1). The new functions computed in the interface of $t$ with $t+1$ are then used by $t+1$, when we perform message passing in $t+1$.

Three important technical issues remain to be discussed. First, message-passing requires the operations of multiplication and marginalization to be performed on functions in each node. These operators can be constructed for HDMNs in a straightforward way by combining the operators by [Lauritzen, 1992] and [Larkin and Dechter, 2003] that work on HBNs and discrete mixed networks respectively. We will now briefly comment on how the multiplication operator can be derived. Let us assume we want to multiply a collection of probabilistic functions $P'$ and a set of constraint relations $C'$ (which consist of only discrete tuples allowed by the constraint) to form a single function $PC$. Here, multiplication can be performed on the functions in $P'$ and $C'$ separately using the operators in [Lauritzen, 1992] and [Dechter, 2003] respectively to compute a single probabilistic function $P$ and a single constraint relation $C$. These two functions $P$ and $C$ can be multiplied by deleting all tuples in $P$ that are not present in $C$ to form the required function $PC$.

Second, because IJGP(i)-S constructs join-graphs sequentially, the maximum-$i$-bound for IJGP(i)-S is bounded by the treewidth of the time slice and its interfaces and not the treewidth of the entire HDMN model (see Figure 1).

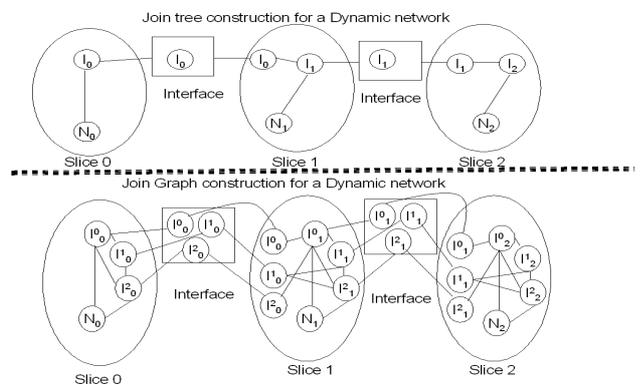

Figure 1: Schematic illustration of the Procedure used for creating join-graphs and join-trees of HDMNs

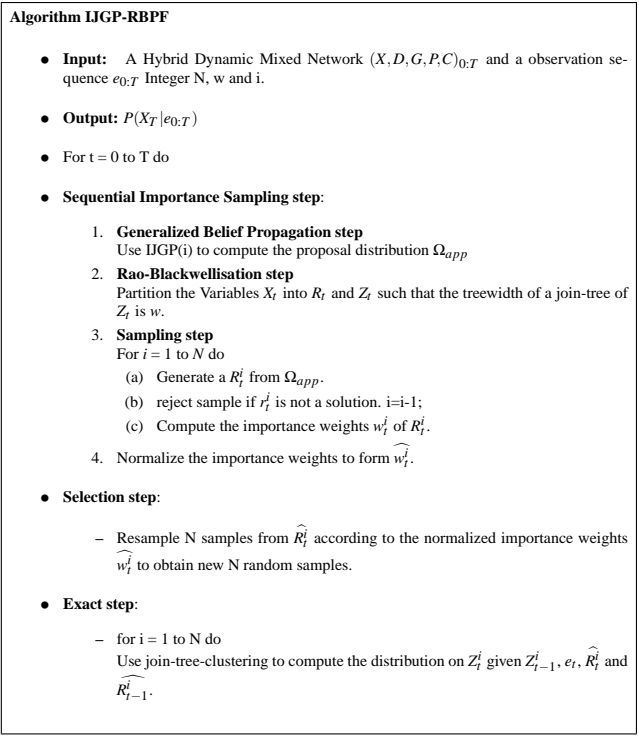

Figure 2: IJGP-RBPF for HDMNs

Third, IJGP(i) guarantees that the computations will be exact if $i$ is equal to the treewidth. This is not true for IJGP(i)-S in general as shown in [Lerner, 2002]. It can be proved that:

THEOREM **3.1** *The complexity of IJGP(i)-S is $O(((|\Delta_t| + n) * d^i * \Gamma_t^3) * T)$ where $|\Delta_t|$ is the number of discrete variables in time-slice t, d is the maximum-domain size of the discrete variables, i is the i-bound used, n is the number of nodes in a join-graph of the time-slice, $\Gamma_t$ is the maximum number of continuous variables in the clique of the join-graph used and T is the number of time-slices.*

## 4 RAO-BLACKWELLISED PARTICLE FILTERING

In this section, we will extend the Rao-Blackwellised Particle filtering algorithm [Doucet et al., 2000] from HDBNs to HDMNs. Before, we present this extension, we will briefly review Particle Filtering and Rao-Blackwellised Particle Filtering (RBPF) for HDBNs.

Particle filtering uses a weighted set of samples or particles to approximate the filtering distribution. Thus, given a set of particles $X_t^1, \ldots, X_t^N$ approximately distributed according to the target-filtering distribution $P(X_t = M|e_{0:t})$, the filtering distribution is given by $P(X_t = M|e_{0:t}) = 1/N \sum_{i=1}^{N} \delta(X_t^i = M)$ where $\delta$ is the Dirac-delta function. Since we cannot sample from $P(X_t = M|e_{0:t})$ directly, Particle filtering uses an appropriate (importance) proposal distribution $Q(X)$ to sample from. The particle filter starts by generating $N$ particles according to an initial proposal distribution $Q(X_0|e_0)$. At each step, it generates the next state $X_{t+1}^i$ for each particle $X_t^i$ by sampling from $Q(X_{t+1}|X_t^i, e_{0:t})$. It then computes the weight of each particle based given by $w_t = P(X)/Q(X)$ to compute a weighted distribution and then *re-samples* from the weighted distribution to obtain a set of un-biased or un-weighted particles.

Particle filtering often shows poor performance in high-dimensional spaces and its performance can be improved by sampling from a sub-space by using the *Rao-Blackwellisation (RB) theorem* (and the particle filtering is called Rao-Blackwellised Particle Filtering (RBPF)). Specifically, the state $X_t$ is divided into two-sets: $R_t$ and $Z_t$ such that only variables in set $R_t$ are sampled (from a proposal distribution $Q(R_t)$) while the distribution on $Z_t$ is computed analytically given each sample on $R_t$ (assuming that $P(Z_t|R_t, e_{0:t}, R_{t-1})$ is tractable). The complexity of RBPF is proportional to the complexity of exact inference step i.e. computing $P(Z_t|R_t, e_{0:t}, R_{t-1})$ for each sample $R_t^k$. w-cutset [Bidyuk and Dechter, 2004] is a parameterized way to select $R_t$ such that the complexity of computing $P(Z_t|R_t, e_{0:t}, R_{t-1})$ is bounded exponentially by $w$. Below, we use the $w$-cutset idea to perform RBPF in HDMNs.

Since exact inference can be done in polynomial time if a HDBN contains only continuous variables, a straightforward application of RBPF to HDBNs involves sampling only the discrete variables in each time slice and exactly inferring the continuous variables [Lerner, 2002].

Extending this idea to HDMNs, suggests that in each time slice $t$ we sample the discrete variables and discard all particles that violate the constraints in the time slice. Let us assume that we select a proposal distribution $Q$ that is a good approximation of the probabilistic filtering distribution but ignores the constraint portion. The extension described above can be inefficient because if the proposal distribution $Q$ is such that it makes non-solutions to the constraint portion highly probable, most samples from $Q$ will be rejected (because these samples $R_t^i$ will have $P(R_t^i) = 0$ and so the weight will be zero). Thus, on one extreme sampling only from the Bayesian Network portion of each time-slice may lead to potentially high rejection-rate.

On the other extreme, if we want to make the sample rejection rate zero we would have to use a proposal distribution $Q'$ such that all samples from this distribution are solutions. One way to find this proposal distribution is to make the constraint network backtrack-free (using adaptive-consistency [Dechter, 2003] or exact constraint propagation) along an ordering of variables and then sample along the reverse ordering. Another approach is to use join-tree-clustering which combines probabilistic and deterministic information and then sample from the join-tree. However, both join-tree-clustering and adaptive-consistency are time and space exponential in treewidth and so they are costly when the treewidth is large. Thus on one hand, zero-rejection rate implies using a potentially costly inference procedure while on the other hand sampling from a proposal distribution that ignores constraints may result in a high rejection rate.

We propose to exploit the middle ground between the two extremes by combining the constraint network and the Bayesian Network into a single approximate distribution $\Omega_{app}$ using IJGP(i) which is a bounded inference procedure. Note that because IJGP(i) has polynomial time complexity for constant $i$, we would not eliminate the sample-rejection rate completely. However, by using IJGP(i) we are more likely to reduce the rejection-rate because IJGP(i) also achieves Constraint Propagation and it is well known that Constraint Propagation removes many inconsistent tuples thereby reducing the chance of sampling a non-solution. [Dechter, 2003].

Another important advantage of using IJGP(i) is that it yields very good approximations to the true posterior [Dechter et al., 2002] thereby proving to be an ideal candidate for proposal distribution. Note that IJGP(i) can be used as a proposal distribution because it can be proved using results from [Dechter and Mateescu, 2003] that IJGP(i) includes all supports of $P(X_t|e_{0:t}, X_{t-1})$ (i.e. $P(X_t|e_{0:t}, X_{t-1}) > 0$ implies that the output of IJGP(i) viz. $Q > 0$)

Note that IJGP(i) we use here is different from the algorithm IJGP(i)-S that we described in the previous section. This is because in our RBPF procedure, we need to compute an approximation to the distribution $P(R_t|R_{t-1}^k, e_{0:t})$ given the sample $R_{t-1}^k$ on variables $R_{t-1}$ and evidence $e_{0:t}$. IJGP(i) as used in our RBPF procedure works on HMNs and can be derived using the results in [Dechter et al., 2002, Lauritzen, 1992, Larkin and Dechter, 2003]. For lack of space we do not describe the details of this algorithm (see a companion paper [Gogate and Dechter, 2005] for details).

The integration of the ideas described above into a formal algorithm called IJGP-RBPF is given in Figure 2. It uses the same template as in [Doucet et al., 2000] and the only step different in IJGP-RBPF from the original template is the implementation of the Sequential Importance Sampling step (SIS).

SIS is divided into three-steps: (1) In the Generalized Belief Propagation step of SIS, we first perform Belief Propagation using IJGP(i) to form an approximation of the posterior (say $\Omega_{app}$) as described above. (2) In the Rao-Blackwellisation step, we first partition the variables in a 2THMN into two sets $R_t$ and $Z_t$ using a method due to [Bidyuk and Dechter, 2004]. This method [Bidyuk and Dechter, 2004] removes minimal variables $R_t$ from $X_t$ such that the treewidth of the remain-ing network $Z_t$ is bounded by $w$. (3) In the sampling step, the variables $R_t$ are sampled from $\Omega_{app}$. To generate a sample from $\Omega_{app}$, we use a special data-structure of ordered buckets which is described in a companion paper [Gogate and Dechter, 2005]. Importance weights are computed as usual [Doucet et al., 2000].

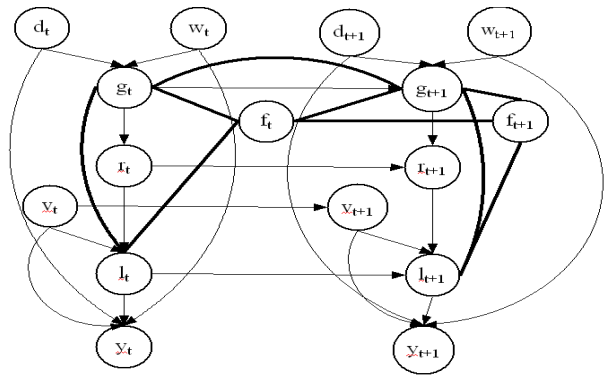

Figure 3: Car Travel Activity model of an individual

Finally, the exact-step computes a distribution on $Z_t$ using join-tree-clustering for HMNs (see a companion paper [Gogate and Dechter, 2005] for details on join-tree-clustering for HMNs). It can be proved that:

THEOREM **4.1** *The complexity of IJGP-RBPF(i,w) is $O([N_R * d^{w+1} + ((|\Delta| + n) * (d^i * |\Gamma|^3))] * T)$ where $|\Delta|$ is the number of discrete variables, d is the maximum-domain size of the discrete variables, i is the adjusted-i-bound, w is defined by w-cutset, n is the number of nodes in a join-graph, $\Gamma$ is the number of continuous variables in a 2-THMN, $N_R$ is the number of samples actually drawn and T is the number of time-slices.*

## 5 THE TRANSPORTATION MODEL

In this section, we describe the application of HDMNs to a real-world problem of inferring car travel activity of individuals. The major query in our HDMN model is to predict where a traveler is likely to go and what his/her route to the destination is likely to be, given the current location of the traveler's car. This application was described in [Liao et al., 2004] in a different context for detecting abnormal behavior in Alzheimer's patients and they use a Abstract Hierarchical Markov Models (AHMM) for reasoning about this problem. The novelty in our approach is not only a more general modeling framework and approximate inference algorithms but also a domain independent implementation which allows an expert to add and test variants of the model.

Figure 3 shows a HDMN model for modeling the car travel activity of individuals. Note that the directed links express the probabilistic relationships while the undirected (bold) edges express the constraints.

We consider the roads as a Graph $G(V,E)$ where the vertices $V$ correspond to intersections while the edges $E$ correspond to segments of roads between intersections. The variables in the model are as follows. The variables $d_t$ and $w_t$ represent the information about time-of-day and day-of-week respectively. $d_t$ is a discrete variable and has four values (*morning, afternoon, evening, night*) while the variable $w_t$ has two values (*weekend, weekday*). Variable $g_t$ represents the persons next goal (e.g. his office, home etc). We consider a location where the person spends significant amount of time as a proxy for a goal [Liao et al., 2004]. These locations are determined through a preprocessing step by noting the locations in which the dwell-time is greater than a threshold (15 minutes). Once such locations are determined, we cluster those that are in close proximity to simplify the goal set. A goal can be thought of as a set of edges $E_1 \subset E$ in our graph representation. The route level $r_t$ represents the route taken by the person to move from one goal to other. We arbitrarily set the number of values it can take to $|g_t|^2$. The person's location $l_t$ and velocity $v_t$ are estimated from GPS reading $y_t$. $f_t$ is a counter (essentially goal duration) that governs goal switching. The Location $l_t$ is represented in the form of a two-tuple $(a,w)$ where $a = (s_1, s_2), a \in E$ and $s_1, s_2 \in V$ is an edge of the map $G(V,E)$ and $w$ is a Gaussian whose mean is equal to the distance between the person's current position on $a$ and one of the intersections in $a$.

The probabilistic dependencies in the model are straightforward and can be found by tracing the arrows (see Figure 3). The constraints in the model are as follows. We assume that a person switches his goal from one time slice to another when he is near a goal or moving away from a goal but not when he is on a goal location. We also allow a forced switch of goals when a specified maximum time that he is supposed to spend at a goal is reached. This is modeled by using a constant D. These assumptions of switching between goals is modeled using the following constraints between the current location, the current goal, the next goal and the switching counters: (1)If $l_{t-1} = g_{t-1}$ and $F_{t-1} = 0$ Then $F_t = D$, (2) If $l_{t-1} = g_{t-1}$ and $F_{t-1} > 0$ Then $F_t = F_{t-1} - 1$, (3) If $l_{t-1} \neq g_{t-1}$ and $F_{t-1} = 0$ Then $F_t = 0$ and (4) If $l_{t-1} \neq g_{t-1}$ and $F_{t-1} > 0$ Then $F_t = 0$, (5) If $F_{t-1} > 0$ and $F_t = 0$ Then $g_t$ is given by $P(g_t|g_{t-1})$, (6) If $F_{t-1} = 0$ and $F_t = 0$ Then $g_t$ is same as $g_{t-1}$, (7) If $F_{t-1} > 0$ and $F_t > 0$ $g_t$ is same as $g_{t-1}$ and (8) If $F_{t-1} = 0$ and $F_t > 0$ $g_t$ is given by $P(g_t|g_{t-1})$.

# 6 EXPERIMENTAL RESULTS

The test data consists of a log of GPS readings collected by one of the authors. The test data was collected over a six month period at intervals of 1-5 seconds each. The data consist of the current time, date, location and velocity of the person's travel. The location is given as latitude and longitude pairs. The data was first divided into individual routes taken by the person and the HDMN model was learned using the Monte Carlo version of the EM algorithm [Liao et al., 2004, Levine and Casella, 2001].

We used the first three months' data as our training set while the remaining data was used as a test set. TIGER/Line files available from the US Census Bureau formed the graph on which the data was snapped. As specified earlier our aim is two-fold: (a) Finding the destination or goal of a person given the current location and (b) Finding the route taken by the person towards the destination or goal.

To compare our inference and learning algorithms, we use three HDMN models. Model-1 is the model shown in Figure 3. Model-2 is the model given in Figure 3 with the variables $w_t$ and $d_t$ removed from each time-slice. Model-3 is the base-model which tracks the person without any high-level information and is constructed from Figure 3 by removing the variables $w_t$, $d_t$, $f_t$, $g_t$ and $r_t$ from each time-slice.

We used 4 inference algorithms. Since EM-learning uses inference as a sub-step, we have 4 different learning algorithms. We call these algorithms as IJGP-S(1), IJGP-S(2) and IJGP-RBPF(1,1,N) and IJGP-RBPF(1,2,N) respectively. Note that the algorithm IJGP-S(i) (described in Section 3) uses $i$ as the $i$-bound. IJGP-RBPF(i,w,N) (described in Section 4) uses $i$ as the $i$-bound for IJGP(i), $w$ as the $w$-cutset bound and $N$ is the number of particles at each time slice. Three values of $N$ were used: 100, 200 and 500. For EM-learning, $N$ was 500. Experiments were run on a Pentium-4 2.4 GHz machine with 2G of RAM. Note that for Model-1, we only use IJGP-RBPF(1,1) and IJGP(1)-S because the maximum $i$-bound in this model is bounded by 1 (see section 3).

## 6.1 FINDING DESTINATION OR GOAL OF A PERSON

The results for goal prediction with various combinations of models, learning and inference algorithms are shown in Tables 1, 2 and 3. We define prediction accuracy as the number of goals predicted correctly. Learning was performed offline. Our slowest learning algorithm based on GBP-RBPF(1,2) used almost 5 days of CPU time for Model-1, and almost 4 days for Model-2—significantly less than the period over which the data was collected. The column 'Time' in Tables 1, 2 and 3 shows the time for inference algorithms in seconds while the other entries indicate the accuracy for each combination of inference and learning algorithms.

In terms of which model yields the best accuracy, we can see that Model-1 achieves the highest prediction accuracy of 84% while Model-2 and Model-3 achieve prediction accuracies of 77% and 68% respectively or lower.

For Model-1, to verify which algorithm yields the best learned model we see that IJGP-RBPF(1,2) and IJGP(2)-S yield an accuracy of 83% and 81% respectively while for Model-2, we see that the average accuracy of IJGP-RBPF(1,2) and IJGP(2)-S was 76% and 75% respectively. From these two results, we can see that IJGP-RBPF(1,2) and IJGP(2)-S are the best performing learning algorithms.

For Model-1 and Model-2, to verify which algorithm yields the best accuracy given a learned model, we see that IJGP(2)-S is the most cost-effective alternative in terms time versus accuracy while IJGP-RBPF yields the best accuracy.

Table 1: Goal-prediction: Model-1

|   |   |   | LEARNING | | | |
|---|---|---|---|---|---|---|
|   |   |   | IJGP-RBPF | | IJGP-S | |
| N | Inference | Time | (1,1) | (1,2) | (1) | (2) |
| 100 | IJGP-RBPF(1,1) | 12.3 | 78 | 80 | 79 | 80 |
| 100 | IJGP-RBPF(1,2) | 15.8 | 81 | 84 | 78 | 81 |
| 200 | IJGP-RBPF(1,1) | 33.2 | 80 | 84 | 77 | 82 |
| 200 | IJGP-RBPF(1,2) | 60.3 | 80 | 84 | 76 | 82 |
| 500 | IJGP-RBPF(1,1) | 123.4 | 81 | 84 | 80 | 82 |
| 500 | IJGP-RBPF(1,2) | 200.12 | 84 | 84 | 81 | 82 |
|   | IJGP(1)-S | 9 | 79 | 79 | 77 | 79 |
|   | IJGP(2)-S | 34.3 | 74 | 84 | 78 | 82 |
|   | Average |   | 79.625 | 82.875 | 78.25 | 81.25 |

## 6.2 FINDING THE ROUTE TAKEN BY THE PERSON

To see how our models predict a person's route, we use the following method. We first run our inference algorithm on the learned model and predict the route that the person is likely to take. We then super-impose this route on the actual route taken by the person. We then count the number of roads that were not taken by the person but were in the predicted route i.e. the false positives, and also compute the number of roads that were taken by the person but were not in the actual route i.e. the false negatives. The two measures are reported in Table 4 for the best performing learning models in each category: viz GBP-RBPF(1,2) for Model-1 and Model-2 and GBP-RBPF(1,1) for Model-3. As we can see Model-1 and Model-2 have the best route prediction accuracy (given by low false positives and false negatives).

Table 2: Goal Prediction: Model-2

|   |   |   | LEARNING | | | |
|---|---|---|---|---|---|---|
|   |   |   | IJGP-RBPF | | IJGP-S | |
| N | Inference | Time | (1,1) | (1,2) | (1) | (2) |
| 100 | IJGP-RBPF(1,1) | 8.3 | 73 | 73 | 71 | 73 |
| 100 | IJGP-RBPF(1,2) | 14.5 | 76 | 76 | 71 | 75 |
| 200 | IJGP-RBPF(1,1) | 23.4 | 76 | 77 | 71 | 75 |
| 200 | IJGP-RBPF(1,2) | 31.4 | 76 | 77 | 71 | 76 |
| 500 | IJGP-RBPF(1,1) | 40.08 | 76 | 77 | 71 | 76 |
| 500 | IJGP-RBPF(1,2) | 51.87 | 76 | 77 | 71 | 76 |
|   | IJGP(1)-S | 6.34 | 71 | 73 | 71 | 74 |
|   | IJGP(2)-S | 10.78 | 76 | 76 | 72 | 76 |
|   | Average |   | 75 | 75.75 | 71.125 | 75.125 |

Table 3: Goal Prediction Model-3

|   |   |   | LEARNING | |
|---|---|---|---|---|
| N | Inference | Time | IJGP-RBPF(1,1) | IJGP(1)-S |
| 100 | IJGP-RBPF(1,1) | 2.2 | 68 | 61 |
| 200 | IJGP-RBPF(1,1) | 4.7 | 67 | 64 |
| 500 | IJGP-RBPF(1,1) | 12.45 | 68 | 63 |
|   | IJGP(1)-S | 1.23 | 66 | 62 |
|   | Average |   | 67.25 | 62.5 |

## 7 RELATED WORK

[Liao et al., 2004] and [Patterson et al., 2003] describe a model based on AHMEM [Bui, 2003] and Hierarchical Markov Models (HMMs) respectively for inferring high-level behavior from GPS-data. Our model goes beyond their model by representing two new variables day-of-week and time-of-day which improves the accuracy in our model by about 6%.

A mixed network framework for representing deterministic and uncertain information was presented in [Dechter and Larkin, 2001, Larkin and Dechter, 2003, Dechter and Mateescu, 2004]. These previous works also describe exact inference algorithms for mixed networks with the restriction that all variables should be discrete. Our work goes beyond these previous works in that we describe approximate inference algorithms for the mixed network framework, allow continuous Gaussian nodes with certain restrictions in the mixed network framework and model discrete-time stochastic processes. The approximate inference algorithms called IJGP(i) described in [Dechter et al., 2002] handled only discrete variables. In our work, we extend this algorithm to include Gaussian variables and discrete constraints. We also develop a sequential version of this algorithm for dynamic models.

Particle Filtering is a very attractive research area [Doucet et al., 2000]. Particle Filtering in HDMNs can be inefficient if non-solutions of constraint portion have high probability of being sampled. We show how to alleviate this difficulty by performing IJGP(i) before sampling. This algorithm IJGP-RBPF yields the best performance in our settings and might prove to be useful in applications in which particle filtering is preferred.

Table 4: False positives (FP) and False negatives for routes taken by a person (FN)

|   |   | Model1 | Model2 | Model3 |
|---|---|---|---|---|
| N | INFERENCE | FP/FN | FP/FN | FP/FN |
|   | IJGP(1) | 33/23 | 39/34 | 60/55 |
|   | IJGP(2) | 31/17 | 39/33 |   |
| 100 | IJGP-RBPF(1,1) | 33/21 | 39/33 | 60/54 |
| 200 | IJGP-RBPF(1,1) | 33/21 | 39/33 | 58/43 |
| 100 | IJGP-RBPF(1,2) | 32/22 | 42/33 |   |
| 200 | IJGP-RBPF(1,2) | 31/22 | 38/33 |   |

## 8 CONCLUSION AND FUTURE WORK

In this paper, we introduced a new modeling framework called HDMNs, a representation that handles discrete-time-stochastic processes, deterministic and probabilistic information on both continuous and discrete variables in a systematic way. We also propose a GBP-based algorithm called IJGP(i)-S for approximate inference in this framework. The main algorithmic contribution of this paper is presenting a class of Rao-Blackwellised particle filtering algorithm, IJGP-RBPF for HDMNs which integrates a generalized belief propagation component with a Rao-Blackwellised Particle Filtering scheme for effective sampling in the presence of constraints. Another contribution of this paper is addressing a complex and highly relevant real life domain using a general framework and domain independent algorithms. Directions for future work include relaxing the restrictions made on dependencies between discrete and continuous variables and developing an efficient EM-algorithm.

## ACKNOWLEDGEMENTS

The first and third author were supported in part by National Science Foundation under award numbers 0331707 and 0331690. The second-author was supported in part by the NSF grant IIS-0412854.